\documentclass[5p,10pt,number]{elsarticle}

\usepackage{amsmath,amssymb}
\usepackage{newtxtext,newtxmath}
\usepackage{graphicx}
\usepackage{booktabs}
\usepackage{array}
\usepackage{multirow}
\usepackage{float}
\usepackage{algorithm}
\usepackage{algorithmic}
\usepackage[font=footnotesize]{caption}
\usepackage{setspace}
\usepackage[hidelinks]{hyperref}

\journal{Pattern Recognition}
\biboptions{sort&compress}

\begin{document}
\singlespacing

\begin{frontmatter}

\title{MSCGC-KAN: Multi-scale Causal Graph Convolution and KAN-inspired Analytic-basis Mapping for EEG Emotion Recognition}

\author[inst1]{Haoliang Gong}
\author[inst1,inst2]{Qingshan She\corref{cor1}}
\ead{qsshe@hdu.edu.cn}
\author[inst1]{Jiale Xu}
\author[inst1,inst2]{Yunyuan Gao}
\author[inst1,inst2]{Xugang Xi}
\cortext[cor1]{Corresponding author: Qingshan She}
\address[inst1]{School of Automation, Hangzhou Dianzi University, Hangzhou 310018, China}
\address[inst2]{Zhejiang Provincial Key Laboratory of Brain Computer Collaborative Intelligence Technology and Applications}

\begin{abstract}
Electroencephalogram (EEG)-based emotion recognition is an important affective computing task, and recent EEG foundation models provide useful generic representations for downstream adaptation. However, simple downstream heads may not fully exploit pretrained EEG representations because they provide insufficient modeling of multi-scale temporal dynamics and task-driven inter-channel dependencies and rely on linear classification. To address these issues, this paper proposes a new EEG emotion recognition method, termed MSCGC-KAN, which introduces a structured task head composed of multi-scale causal graph convolution and KANLinear, a KAN-inspired analytic-basis mapping module. Built on a pre-trained CBraMod backbone, MSCGC-KAN enhances downstream adaptation by jointly strengthening multi-scale temporal modeling, task-driven inter-channel dependency modeling, and nonlinear discriminative mapping within a compact task-specific head. This design preserves the representation advantage of the foundation model while making the classifier more sensitive to emotion-related spatiotemporal patterns. Experiments were conducted on the public FACED and SEED-VII datasets. The proposed method achieves a balanced accuracy of 60.66\%, a Cohen's Kappa of 0.5525, and a weighted F1-score of 60.40\% on FACED, and obtains 33.27\%, 0.2223, and 33.64\%, respectively, on SEED-VII. Compared with the CBraMod+Linear baseline, the balanced accuracy is improved by 5.91 and 2.03 percentage points on the two datasets, respectively. Under their respective subject-disjoint and subject-dependent evaluation protocols, these results indicate that structured task-head design is an effective way to improve EEG emotion recognition when fine-tuning pre-trained EEG models.
\end{abstract}


\end{frontmatter}

\section{Introduction}

Emotion recognition enables intelligent systems to infer affective states and supports human--computer interaction, healthcare, mental-state monitoring, and brain--computer interfaces \cite{ref1}. Common modalities include facial expressions \cite{ref2}, speech \cite{ref3}, and text \cite{ref4}; however, these observable signals can be affected by voluntary masking, environmental conditions, and individual expression habits.

Physiological signals more directly reflect internal affective responses. EEG is particularly attractive because it provides noninvasive neural measurements with high temporal resolution, but its weak, non-stationary, and noisy signals vary substantially across subjects. These properties complicate the extraction and transfer of stable emotion representations \cite{ref5}, despite evidence that EEG contains temporally persistent affective patterns \cite{ref6}.

EEG emotion recognition has consequently progressed from expert-designed features to learned representations, while cross-subject generalization, robustness, and interpretability remain open challenges \cite{ref27}. Recent temporal models combine multi-anchor temporal and space-aware convolution \cite{ref7}, three-dimensional representations with dilated fully convolutional networks \cite{ref19}, or multi-scale dynamic convolution with gated Transformers and temporal convolution \cite{ref28}. These advances capture local and global dynamics, but temporal aggregation alone does not explicitly integrate channel organization, spectral variation, and model-level inspection of downstream discrimination.

Complementary approaches jointly encode spatial, spectral, and temporal information through three-dimensional dense attention \cite{ref9}, combine channel and self-attention \cite{ref10}, or use Transformers and representation visualization to support model interpretation \cite{ref33}. Collectively, they show that affective EEG patterns depend on coordinated variation across multiple dimensions.

Graph neural networks explicitly model relations among EEG channels. Existing designs use sparse dynamic graphs \cite{ref11}, regularized relation learning \cite{ref12}, sample-adaptive adjacency matrices \cite{ref13}, channel--frequency dual attention \cite{ref29}, temporal-frequency graphs with cross-subject alignment \cite{ref30}, and directed sparse graphs with attention \cite{ref14}. Although these methods demonstrate the value of structured channel-dependency modeling, its integration with downstream classifier design during foundation-model fine-tuning remains underexplored.

Self-supervised learning has also produced transferable EEG foundation models, including CBraMod's criss-cross region modeling \cite{ref18}, BIOT's cross-dataset biosignal Transformer \cite{ref20}, large-scale brain representations for BCI \cite{ref21}, and graph-enhanced pre-training \cite{ref32}. However, this literature primarily develops the backbone, leaving downstream task-head design comparatively underexplored. Average pooling followed by a linear projection may not fully exploit the multi-scale temporal, inter-channel, and nonlinear structure of pre-trained features. This motivates a compact task head that jointly reorganizes spatiotemporal information and provides an analyzable nonlinear mapping for emotion discrimination.

Recent KANs model nonlinear transformations using learnable univariate functions on network edges, commonly parameterized by splines \cite{ref17}. Our module does not implement a conventional KAN and is not presented as a realization of the Kolmogorov--Arnold representation theorem. Instead, it adopts fixed analytic transformations as a lightweight nonlinear feature-expansion mechanism. We retain KANLinear as the implementation-level module name and characterize its computation as KAN-inspired analytic-basis mapping. Here, KAN-inspired refers only to the limited design cue of applying univariate nonlinear transformations and does not imply architectural equivalence to a spline-based KAN.

We therefore propose MSCGC-KAN, a structured task head for fine-tuning EEG foundation models. After the pre-trained backbone, multi-scale causal residual graph convolution captures temporal dynamics at heterogeneous ranges and learns task-driven dependencies among channel representations. The KANLinear analytic-basis mapping then replaces a single linear projection with lightweight nonlinear feature expansion. This design preserves generic pre-trained representations while adapting them to emotion-specific discrimination.

The main contributions of this work are summarized as follows.
\begin{enumerate}
\item A structured task-head framework is proposed for EEG foundation model fine-tuning. Without heavily modifying the backbone, the framework unifies multi-scale temporal modeling, task-driven inter-channel dependency modeling, and nonlinear discriminative enhancement within a single downstream classification head.
\item A multi-scale causal residual graph convolution module is designed to capture temporal dynamics at different scales using causal padding as a directional temporal inductive bias and to model task-driven inter-channel dependencies through a learnable graph structure.
\item A lightweight KANLinear module, formulated as a KAN-inspired analytic-basis mapping, is constructed by replacing a single linear projection with fixed analytic basis expansion, thereby improving nonlinear discrimination of high-dimensional EEG features under a manageable parameter cost.
\item Experiments on the public FACED and SEED-VII datasets, together with ablation studies and visualization analyses, verify the effectiveness of the proposed method while providing qualitative model-level insight into its discriminative behavior.
\end{enumerate}

\section{Methods}

\begin{figure*}[!t]
\centering
\includegraphics[width=0.96\textwidth]{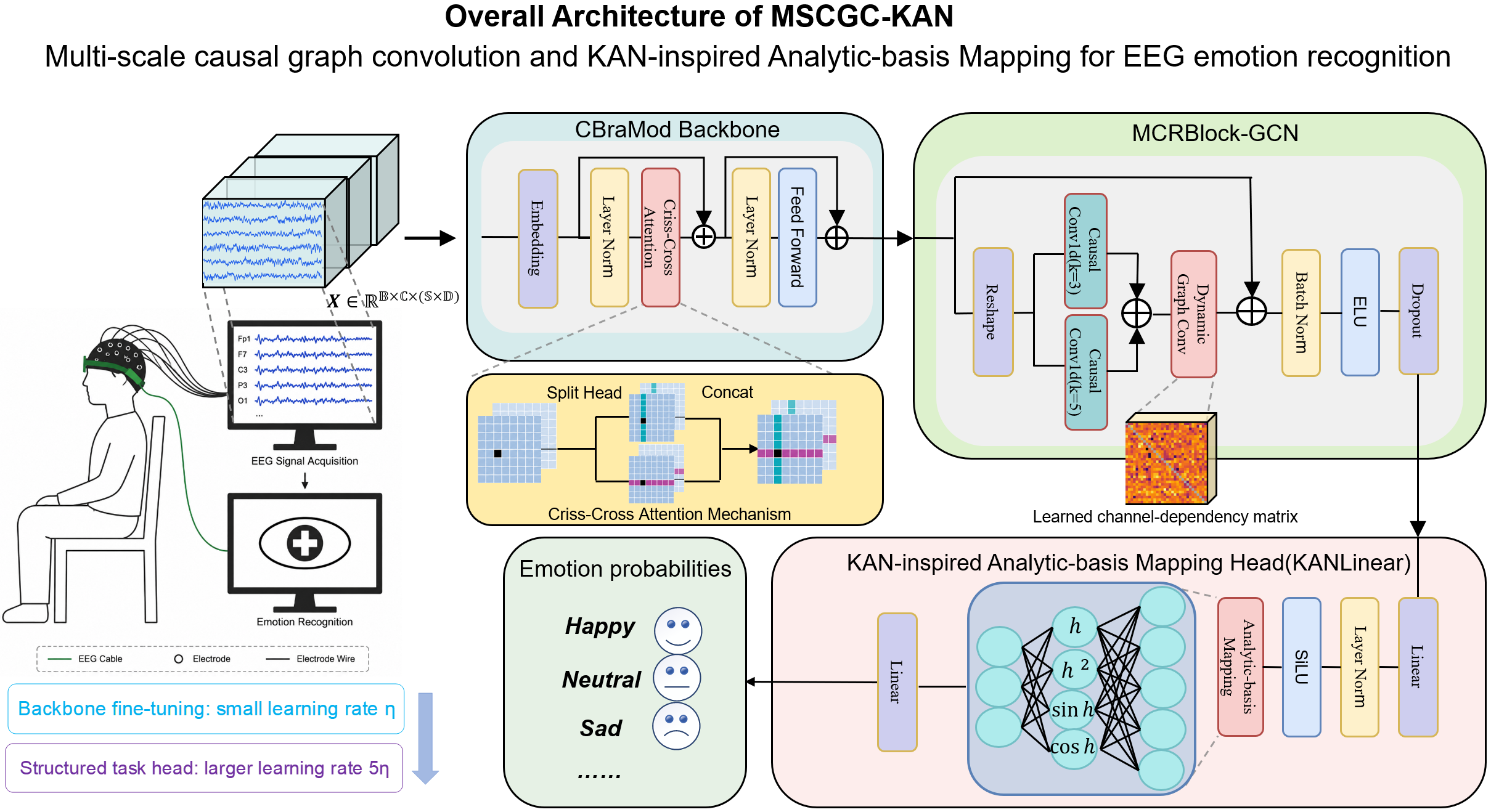}
\caption{Overall architecture of the proposed MSCGC-KAN model. CBraMod first extracts generic EEG representations, MCRBlock-GCN then performs multi-scale causal temporal modeling and learnable graph-based spatial modeling, and KANLinear finally maps the enhanced features to an expressive nonlinear discriminative space for emotion classification.}
\label{fig:mscgckan_architecture}
\end{figure*}

\subsection{Overall Framework}

This article proposes a structured task-head framework for EEG emotion recognition, termed MSCGC-KAN, as illustrated in Fig.~\ref{fig:mscgckan_architecture}. The framework is designed to adapt pre-trained EEG representations to emotion recognition by addressing three key requirements: multi-scale temporal modeling, task-driven inter-channel dependency modeling, and inspectable nonlinear feature mapping. Instead of modifying the pre-trained backbone extensively, MSCGC-KAN places a compact enhancement head after CBraMod so that the generic representation ability of the foundation model can be preserved while task-specific discriminative information is further extracted.

The proposed framework consists of three main components: the pre-trained CBraMod backbone, the multi-scale causal residual graph convolution module, and the KANLinear head. CBraMod first serves as a generic EEG encoder and produces high-dimensional features from the input temporal windows. These features are then reorganized into a tensor form suitable for structured downstream modeling. MCRBlock-GCN subsequently performs multi-scale causal temporal modeling and learnable graph-based spatial modeling, enabling the task head to capture both emotion-related temporal dynamics and task-driven inter-channel dependencies. Finally, KANLinear expands the enhanced representation with fixed analytic basis functions before classification, thereby improving nonlinear discrimination under a controlled parameter budget. Some frequently used notations are summarized in Table~\ref{tab:notations} to facilitate reading.

\begin{table*}[!t]
\caption{Description of frequently used notations}
\label{tab:notations}
\centering
\small
\begin{spacing}{1.0}
\setlength{\tabcolsep}{3.5pt}
\renewcommand{\arraystretch}{0.95}
\begin{tabular}{>{\raggedright\arraybackslash}p{0.11\textwidth}>{\raggedright\arraybackslash}p{0.33\textwidth}>{\raggedright\arraybackslash}p{0.11\textwidth}>{\raggedright\arraybackslash}p{0.35\textwidth}}
\toprule
Notation & Description & Notation & Description \\
\midrule
$B$ & Batch size. & $R(k,d)$ & Receptive field for kernel $k$ and dilation $d$. \\
$C$ & Number of EEG channels. & $k,d,t$ & Kernel size, dilation, and temporal index. \\
$S$ & Number of temporal windows. & $d_{\mathrm{in}}$ & KANLinear input dimension. \\
$D$ & Temporal-window feature dimension. & $\boldsymbol{u}$ & Flattened KANLinear input. \\
$\boldsymbol{X}$ & Input EEG mini-batch tensor. & $\boldsymbol{W}_{\mathrm{in}},\boldsymbol{b}_{\mathrm{in}}$ & KANLinear input-projection parameters. \\
$\boldsymbol{F}$ & CBraMod representation. & $\boldsymbol{h},h_j$ & Hidden vector and its $j$th coordinate. \\
$\boldsymbol{H}$ & MCRBlock-GCN representation. & $\phi_{oj}$ & Analytic mapping from $h_j$ to $z_o$. \\
$\boldsymbol{x}$ & MCRBlock-GCN input. & $a_{oj}^{(q)},b_o$ & Basis coefficient and output bias. \\
$\mathcal{K}$ & Set of temporal kernel sizes. & $z_o$ & $o$th KANLinear output coordinate. \\
$\boldsymbol{o}_{k}$ & Output of the temporal branch with kernel $k$. & $\hat{\boldsymbol{y}}$ & Predicted label distribution. \\
$\boldsymbol{o}$ & Sum of temporal-branch outputs. & $\boldsymbol{I},\boldsymbol{D}$ & Identity and graph-degree matrices. \\
$\boldsymbol{A}$ & Learned inter-channel dependency matrix. & $\boldsymbol{Z}$ & Graph-propagated representation. \\
$\tilde{\boldsymbol{A}},\hat{\boldsymbol{A}}$ & Self-loop-augmented and normalized dependency matrices. & $j,o,q$ & Hidden, output, and basis indices, with $q\in\{1,2,3,4\}$. \\
\bottomrule
\end{tabular}
\end{spacing}
\end{table*}

This design follows the principle that the foundation model provides robust generic representations, while the task head performs structured task-specific enhancement. In this way, intrusive changes to the pre-trained backbone are avoided, and the effect of downstream adaptation can be studied more clearly. For clarity, the detailed computational flow is described as follows. Given an EEG mini-batch $\boldsymbol{X}\in\mathbb{R}^{B\times C\times S\times D}$, where $B$ is the batch size, $C$ is the number of channels, $S$ is the number of temporal windows, and $D$ is the feature dimension of each temporal window, CBraMod first obtains a generic EEG representation. The backbone output is then reshaped and forwarded to MCRBlock-GCN for joint enhancement along the temporal and channel dimensions. Finally, the resulting features are flattened and mapped by KANLinear followed by a linear classifier. The processing pipeline can be written as
\begin{equation}
\boldsymbol{F}=\operatorname{CBraMod}(\boldsymbol{X}),
\end{equation}
\begin{equation}
\boldsymbol{H}=\operatorname{MCRBlock\text{-}GCN}(\operatorname{Reshape}(\boldsymbol{F})),
\end{equation}
\begin{equation}
\hat{\boldsymbol{y}}=\operatorname{Classifier}(\operatorname{KANLinear}(\operatorname{Flatten}(\boldsymbol{H}))).
\end{equation}

For both datasets, the shape sequence is $\boldsymbol{X},\boldsymbol{F}:(B,C,10,200)\rightarrow\boldsymbol{x},\boldsymbol{o}_{k},\boldsymbol{o},\boldsymbol{Z},\boldsymbol{H}:(B,C,2000)$; the implemented reshape $b\,c\,s\,d\rightarrow b\,c\,(s d)$ merges only $s$ and $d$. For FACED ($C=32$), flattening and the KANLinear hidden/basis/mapped, linear, and logit stages yield $(B,64{,}000)\rightarrow(B,512)\rightarrow(B,2048)\rightarrow(B,2000)\rightarrow(B,200)\rightarrow(B,9)$. For SEED-VII ($C=62$), adaptive pooling first changes $\boldsymbol{H}$ to $(B,62,1000)$, followed by $(B,62{,}000)\rightarrow(B,512)\rightarrow(B,2048)\rightarrow(B,2000)\rightarrow(B,200)\rightarrow(B,7)$. Criss-cross attention contextualizes each token but returns its $(c,s)$ query at the same index, without channel permutation or pooling. Thus, adjacency row and column $c$ follow the original electrode order; nevertheless, each node is contextualized, so $\boldsymbol{A}$ represents task-driven dependencies among channel-indexed representations rather than raw-signal or physiological connectivity.

\subsection{CBraMod Backbone for Feature Encoding}

CBraMod is adopted as the feature-encoding backbone in this work \cite{ref18}.

To clarify data provenance, we used the publicly released checkpoint from the \href{https://huggingface.co/weighting666/CBraMod}{CBraMod public release}. It was pretrained exclusively on the Temple University Hospital EEG Corpus (TUEG), comprising 69,652 clinical EEG recordings from 14,987 subjects across 26,846 sessions. FACED and SEED-VII were used only for downstream fine-tuning and evaluation. No samples, labels, trials, subject identifiers, or dataset-specific statistics from either dataset entered pretraining. Because the anonymized datasets lack cross-dataset identity keys, we state documented dataset- and pipeline-level separation rather than unverifiable natural-person non-overlap.

Within the proposed framework, CBraMod transforms raw EEG samples into high-dimensional spatiotemporal features. Since the backbone already possesses strong generic representation ability after pre-training, this work does not redesign the backbone itself. Instead, the focus is placed on the downstream head, which is expected to further unlock the discriminative potential of the pre-trained representation in the emotion recognition setting.

\subsection{Multi-scale Causal Residual Graph Convolution Module}

Emotion-related EEG patterns typically contain both short-term local fluctuations and long-range dynamic trends. A single receptive field is therefore insufficient to represent the full temporal hierarchy. To address this issue, a multi-scale causal residual graph convolution module, termed MCRBlock-GCN, is designed to perform joint spatiotemporal modeling through the combination of multi-scale causal convolution, graph convolution, and residual fusion.

\subsubsection{Causal temporal modeling}

To model local temporal dependencies at different scales, we employ a multi-branch temporal convolution module with heterogeneous receptive fields. Causal padding is used to impose a directional temporal inductive bias, such that the representation at each temporal position is constructed only from the current and preceding temporal units. Although the complete EEG segment is available in our offline classification setting, this constraint provides a structured form of local temporal modeling and avoids bidirectional temporal mixing within individual convolutional operators. We therefore regard causality as an architectural design choice rather than a strict requirement of the task. Each branch follows the sequence of left-only causal padding, Conv1D, BatchNorm1D, ELU, and Dropout. Let $\boldsymbol{x}$ denote the input sequence of MCRBlock-GCN, which is obtained by reshaping the CBraMod feature. Thus, $\boldsymbol{x}$ is the task-head input representation rather than the raw EEG sample $\boldsymbol{X}$. The $k$-th branch can be written as
\begin{equation}
\boldsymbol{o}_{k}=\operatorname{CausalConv}_{k}(\boldsymbol{x}), \quad k\in\mathcal{K},
\end{equation}
where $\mathcal{K}$ is the set of kernel sizes. For a one-dimensional convolution with kernel size $k$, dilation factor $d$, and stride 1, the receptive field is
\begin{equation}
R(k,d)=1+(k-1)d.
\end{equation}
The standard causal branches used in MCRBlock-GCN have $d=1$, and hence $R(k,1)=k$. With $\mathcal{K}=\{3,5\}$, the two complementary branches aggregate 3 and 5 temporal positions at limited additional branch cost. At position $t$, these ranges are $\{t,t-1,t-2\}$ and $\{t,t-1,t-2,t-3,t-4\}$. The positions index the reshaped CBraMod representation supplied to the task head rather than individual samples of the raw EEG waveform. The outputs of all branches are kept aligned in the channel dimension and fused by summation:
\begin{equation}
\boldsymbol{o}=\sum_{k\in\mathcal{K}}\boldsymbol{o}_{k}.
\end{equation}
Compared with feature concatenation, branch summation enables cross-scale fusion without markedly increasing the parameter count. This parallel multi-scale branch design is also consistent with the classic Inception-style feature extraction idea \cite{ref23}.

\subsubsection{Graph-based spatial dependency modeling}

After temporal enhancement, a learnable graph convolution layer is introduced along the channel dimension to model task-driven dependencies among EEG channel representations. Each EEG channel is treated as a graph node, and a learnable adjacency matrix $\boldsymbol{A}\in\mathbb{R}^{C\times C}$ is used to describe the channel-dependency structure. To stabilize graph construction, an ELU-based transformation is applied to the adjacency parameters, self-loops are added, and the matrix is symmetrically normalized:
\begin{equation}
\tilde{\boldsymbol{A}}=\operatorname{ELU}(\boldsymbol{A})+\boldsymbol{I},
\end{equation}
\begin{equation}
\hat{\boldsymbol{A}}=\boldsymbol{D}^{-\frac{1}{2}}\tilde{\boldsymbol{A}}\boldsymbol{D}^{-\frac{1}{2}},
\end{equation}
where $\boldsymbol{D}$ is the degree matrix. Graph-based feature propagation is then performed as
\begin{equation}
\boldsymbol{Z}=\hat{\boldsymbol{A}}\boldsymbol{o}.
\end{equation}
This operation propagates discriminative information across channel representations. Because $\boldsymbol{A}$ is optimized jointly with the emotion-classification objective, its entries represent task-driven dependency weights that support discrimination. The matrix is not constructed from correlation, coherence, phase-locking value, or another explicit functional-connectivity estimator and therefore should not be interpreted as a direct estimate of physiological connectivity. After training, it can be visualized as a learned channel-dependency matrix or task-driven channel-dependency graph to inspect model behavior.

\subsubsection{Residual enhancement and post-normalization}

To improve training stability and preserve the original semantic information, a residual connection is applied at the end of the module. The graph-enhanced feature is added to the input feature and then post-processed by batch normalization, ELU, and Dropout:
\begin{equation}
\boldsymbol{H}=\operatorname{Dropout}(\operatorname{ELU}(\operatorname{BN}(\boldsymbol{Z}+\boldsymbol{x}))).
\end{equation}
This design helps alleviate gradient degradation while reducing the risk of losing useful discriminative information through excessive graph smoothing. Overall, MCRBlock-GCN follows a ``temporal-first, spatial-second'' pathway and equips the task head with more structured joint spatiotemporal representations.

\subsection{KANLinear: KAN-inspired Analytic-basis Mapping}

Inspired by the use of univariate nonlinear transformations in Kolmogorov--Arnold Networks, rather than implementing a conventional spline-based KAN, we employ a lightweight set of fixed analytic basis functions to enrich the pretrained EEG representation.

\subsubsection{Analytic basis expansion}

Let $\boldsymbol{u}\in\mathbb{R}^{d_{\mathrm{in}}}$ denote the flattened input to the module. The implementation first applies an affine input projection, layer normalization (LN), and the sigmoid linear unit (SiLU) to produce a capacity--cost-balanced 512-dimensional hidden representation
\begin{equation}
\boldsymbol{h}=\operatorname{SiLU}\!\left(\operatorname{LN}\!\left(\boldsymbol{W}_{\mathrm{in}}\boldsymbol{u}+\boldsymbol{b}_{\mathrm{in}}\right)\right).
\end{equation}
For hidden coordinate $h_j$ and output coordinate $z_o$, the learned output projection in KANLinear induces the analytic edge mapping
\begin{equation}
\phi_{oj}(h_j)=a^{(1)}_{oj}h_j+a^{(2)}_{oj}h_j^2+a^{(3)}_{oj}\sin(h_j)+a^{(4)}_{oj}\tanh(h_j),
\end{equation}
The edge-wise contributions are then additively aggregated with the output bias to obtain each output coordinate:
\begin{equation}
z_o=b_o+\sum_{j=1}^{512}\phi_{oj}(h_j).
\end{equation}
The coefficients $a^{(1)}_{oj},\ldots,a^{(4)}_{oj}$ and $b_o$ are learned through the output projection, whereas the four analytic transformations are fixed. In all reported experiments, the implementation uses exactly these four basis functions, spanning linear, polynomial, periodic, and bounded saturating responses. It contains no learnable spline grid or spline coefficients. Accordingly, KANLinear is a KAN-inspired, constrained analytic-basis feature-expansion mechanism rather than a conventional spline-based KAN.

\begin{algorithm}[!t]
\caption{Implementation steps of the proposed MSCGC-KAN}
\label{alg:mscgckan}
\begin{algorithmic}[1]
\REQUIRE Training samples $\{\boldsymbol{X}_{i},y_{i}\}$, pre-trained CBraMod weights, epochs $E$
\ENSURE Best model parameters and test predictions
\STATE Initialize CBraMod and attach MCRBlock-GCN, KANLinear, and a linear classifier.
\STATE Use differential learning rates for the backbone and the task head.
\FOR{epoch $=1$ to $E$}
    \STATE Extract backbone features $\boldsymbol{F}=\operatorname{CBraMod}(\boldsymbol{X})$ for each mini-batch.
    \STATE Obtain $\boldsymbol{H}$ by multi-scale causal convolution, graph convolution, and residual fusion.
    \STATE Map $\boldsymbol{H}$ with KANLinear and predict labels using the final classifier.
    \STATE Optimize cross-entropy loss with gradient clipping and cosine annealing.
    \STATE Save the checkpoint if the validation Kappa is improved.
\ENDFOR
\STATE Load the best checkpoint and report test-set metrics.
\end{algorithmic}
\end{algorithm}

\section{Experiments and Results}

\subsection{Datasets and Experimental Settings}

The proposed method is evaluated on two public EEG emotion recognition datasets, FACED and SEED-VII. FACED is a finer-grained affective computing EEG dataset collected from 123 subjects using 32 electrodes arranged according to the international 10--20 system \cite{ref25}. In the original FACED experiment, each subject watched 28 emotion-elicitation video clips covering nine emotional categories, including four positive emotions, four negative emotions, and neutral emotion. This balanced fine-grained design makes FACED suitable for evaluating multi-class EEG emotion recognition and cross-subject generalization. SEED-VII is a multimodal emotion dataset designed for six basic emotions and neutral emotion \cite{ref45}. It contains EEG and eye-movement recordings from 20 subjects, with four sessions per subject and 20 emotion-eliciting videos per session. In this paper, only the EEG modality of SEED-VII is used. Compared with FACED, SEED-VII uses a higher-density 62-channel EEG configuration and seven emotional categories, making it suitable for evaluating spatial modeling under a different channel scale and label structure. These datasets therefore represent different generalization settings: FACED evaluates subject-disjoint performance, whereas the present SEED-VII protocol evaluates subject-dependent, within-subject-session trial-disjoint performance and does not test unseen-subject generalization.

The preprocessing follows the acquisition and cleaning protocols reported with the two datasets. For FACED, EEG signals were originally recorded at 250 or 1000 Hz and were unified to 250 Hz after acquisition. The official preprocessing selects the last 30 s of each video trial, applies a 0.05--47 Hz band-pass filter, interpolates bad electrodes, removes ocular artifacts by independent component analysis using FP1/FP2 as proxy electro-oculogram channels, and re-references the cleaned signals to the common average reference. Therefore, each preprocessed FACED trial contains $30\times250=7500$ temporal samples. For compatibility with the CBraMod temporal-window input and with the 200 Hz temporal resolution used for SEED-VII, each 30-s FACED trial is further resampled to 200 Hz, resulting in $30\times200=6000$ samples. Each trial is then divided into 30 non-overlapping temporal windows, with 200 samples in each window. To form model inputs while preserving temporal order, the 30-window sequence is further separated into three consecutive 10-window samples, leading to the final input shape $(32,10,200)$.

For SEED-VII, EEG signals were acquired at 1000 Hz using 62 active AgCl electrodes arranged according to the international 10--20 system. The preprocessing described with the dataset first inspects the EEG signals and interpolates bad channels, then applies a 0.1--70 Hz band-pass filter and a 50 Hz notch filter to suppress slow drift, high-frequency noise, and power-line interference. The raw signals are subsequently downsampled to 200 Hz. In this work, the preprocessed EEG sequences are segmented into non-overlapping samples consisting of 10 consecutive temporal windows, with 200 samples in each window. Thus, each SEED-VII input sample covers 2000 temporal samples and has the shape $(62,10,200)$.

\subsection{Data Split Protocol}

FACED uses a fixed subject-disjoint split. Subjects 1--80, 81--100, and 101--123 form the training, validation, and test sets, respectively. This assignment is completed before temporal segmentation. Each 30-window trial is then divided into three consecutive 10-window samples. These derived samples remain grouped with the parent trial. Thus, the segmentation increases the number of model inputs but not the number of independent subjects or trials, and no subject, trial, or derived sample crosses the three subsets. SEED-VII instead uses a subject-dependent, within-subject-session trial-disjoint protocol. For each of the 20 subjects and four sessions, the 20 trials are divided in order into 10 training, 5 validation, and 5 test trials before temporal segmentation. The resulting subsets are then pooled across subjects and sessions. Therefore, the same subject and session contribute distinct trials to all three subsets, but no segment derived from one trial crosses subsets. The SEED-VII results consequently do not represent unseen-subject generalization. For both datasets, preprocessing applies a fixed amplitude scaling factor of $1/100$, rather than normalization statistics estimated from any subset. No data-driven channel selection is performed, and all 32 FACED channels or 62 SEED-VII channels are retained.

\subsection{Evaluation Metrics and Model Selection}

Three metrics are adopted to comprehensively evaluate model performance: balanced accuracy, Cohen's Kappa, and weighted F1-score. Let $M$ denote the number of emotion classes, $N$ the total number of test samples, and $n_{ij}$ the number of samples whose ground-truth label is class $i$ and predicted label is class $j$. For class $c$, true positives, false positives, and false negatives are denoted as $\mathrm{TP}_{c}$, $\mathrm{FP}_{c}$, and $\mathrm{FN}_{c}$, respectively.

Balanced accuracy is defined as the average recall over all classes:
\begin{equation}
\mathrm{BA}=\frac{1}{M}\sum_{c=1}^{M}\frac{\mathrm{TP}_{c}}{\mathrm{TP}_{c}+\mathrm{FN}_{c}}.
\end{equation}
Compared with conventional accuracy, balanced accuracy gives equal importance to each class and is therefore more suitable when class distributions or class-wise recognition difficulties are uneven. In all reported results, accuracy refers to balanced accuracy.

Cohen's Kappa measures the agreement between model predictions and ground-truth labels after discounting chance agreement:
\begin{equation}
\kappa=\frac{p_o-p_e}{1-p_e},
\end{equation}
where
\begin{equation}
p_o=\frac{1}{N}\sum_{i=1}^{M}n_{ii}, \quad
p_e=\frac{1}{N^2}\sum_{i=1}^{M}n_{i\cdot}n_{\cdot i}.
\end{equation}
Here, $p_o$ is the observed agreement, $p_e$ is the expected agreement by chance, $n_{i\cdot}$ is the number of samples belonging to class $i$, and $n_{\cdot i}$ is the number of samples predicted as class $i$. A higher Kappa value indicates better agreement beyond random guessing, making it informative for multi-class emotion recognition.

The weighted F1-score is computed as
\begin{equation}
\mathrm{WF1}=\sum_{c=1}^{M}\frac{n_{c\cdot}}{N}\mathrm{F1}_{c},
\end{equation}
with
\begin{equation}
\begin{aligned}
\mathrm{F1}_{c} &= \frac{2P_cR_c}{P_c+R_c},\\
P_c &= \frac{\mathrm{TP}_{c}}{\mathrm{TP}_{c}+\mathrm{FP}_{c}},\\
R_c &= \frac{\mathrm{TP}_{c}}{\mathrm{TP}_{c}+\mathrm{FN}_{c}},
\end{aligned}
\end{equation}
where $n_{c\cdot}$ is the number of samples belonging to class $c$, and $P_c$ and $R_c$ are the precision and recall of class $c$, respectively. This metric balances precision and recall while weighting each class according to its support.

Model parameters are optimized only on training data. Checkpoints are selected by validation-set Kappa, and the test set is used only after model selection.

\subsection{Training Implementation Details}

To make the experiments reproducible, the main implementation settings used for FACED and SEED-VII are summarized in Table~\ref{tab:training_settings}. All experiments were conducted on an Ubuntu server equipped with an NVIDIA GeForce RTX 3090 GPU. The proposed model was implemented using PyTorch 2.7.0.

\begin{table}[!t]
\caption{Training and Reproducibility Settings}
\label{tab:training_settings}
\centering
\begin{tabular}{p{0.42\columnwidth}p{0.46\columnwidth}}
\toprule
Item & Setting \\
\midrule
GPU & NVIDIA GeForce RTX 3090 \\
Framework & PyTorch 2.7.0 \\
Epochs & 30 \\
Batch size & 64 \\
Optimizer & AdamW \cite{ref24} \\
Weight decay & $5\times10^{-2}$ \\
Backbone learning rate & $1\times10^{-4}$ \\
Task-head learning rate & $5\times10^{-4}$ \\
Gradient clipping & Maximum norm of 1.0 \\
Learning-rate scheduler & Cosine annealing \\
Minimum learning rate & $1\times10^{-6}$ \\
Dropout & 0.1 \\
MCRBlock-GCN kernels & $\mathcal{K}=\{3,5\}$ \\
Independent runs & 5 per directly evaluated configuration (different random seeds) \\
\bottomrule
\end{tabular}
\end{table}

The CBraMod backbone is initialized with pre-trained weights and is not frozen during fine-tuning. The backbone uses a learning rate of $1\times10^{-4}$, whereas MCRBlock-GCN and KANLinear use $5\times10^{-4}$, yielding a $1{:}5$ ratio. This conservatively updates the pretrained backbone while allowing the new task head to adapt faster. All four settings were fixed across the reported experiments without exhaustive dataset-specific or test-set-guided search.

Cross-entropy loss is adopted for both datasets. Gradient clipping is applied after back-propagation to stabilize optimization, and the cosine annealing scheduler is updated at each training iteration to gradually reduce the learning rate during fine-tuning. Together with weight decay and dropout, these strategies help preserve the useful knowledge in the pre-trained backbone while allowing the task head to acquire task-specific discrimination. The checkpoint with the best validation-set Kappa score is selected and then evaluated on the test set.

Each configuration evaluated under our experimental protocol was independently trained and tested five times using different random seeds. The reported mean and standard deviation summarize performance across these five runs and therefore quantify run-to-run variability under random initialization.

For each directly evaluated configuration, a two-sided 95\% confidence interval (CI) for the mean performance was additionally calculated using Student's $t$-distribution as
\begin{equation}
\mathrm{CI}_{95\%}=\bar{m}\pm t_{0.975,n-1}\frac{s}{\sqrt{n}},
\end{equation}
where $\bar{m}$ and $s$ denote the sample mean and standard deviation across the $n=5$ independent runs, respectively, and $t_{0.975,4}=2.776$. The intervals quantify uncertainty in the estimated mean across random seeds and are reported in the FACED and SEED-VII ablation tables.

\subsection{Results on the FACED Dataset}

The FACED comparison includes EEGNet \cite{ref36}, EEGConformer \cite{ref37}, SPaRCNet \cite{ref38}, ContraWR \cite{ref39}, CNN-Transformer \cite{ref40}, FFCL \cite{ref41}, ST-Transformer \cite{ref42}, BIOT \cite{ref20}, and LaBraM-Base \cite{ref21}. The external values come from the FACED evaluation reported with CBraMod \cite{ref18}. Our BIOT and LaBraM reproductions were below the published values. To avoid favoring MSCGC-KAN with lower baselines, the table retains their higher literature results. Because preprocessing, splits, implementation, or model selection may still differ for the external baselines, these rows provide a reference-based comparison rather than a fully controlled test.

Table~\ref{tab:faced_compare} shows that MSCGC-KAN records the highest numerical values in the reference-based FACED comparison, supporting further investigation of structured downstream heads. Because the external baselines were obtained from previously reported evaluations rather than paired runs under our protocol, this numerical ranking is interpreted descriptively and not as evidence of statistical superiority.

\begin{table}[!t]
\renewcommand{\arraystretch}{1.15}
\caption{Performance comparison of different methods on the FACED dataset (mean$\pm$std). The CBraMod+Linear and MSCGC-KAN results summarize five independent runs under our protocol; the remaining baseline values are reproduced from the reference evaluation described in the text.}
\label{tab:faced_compare}
\centering
\resizebox{\columnwidth}{!}{%
\begin{tabular}{lccc}
\toprule
Method & Balanced Accuracy & Kappa & Weighted F1 \\
\midrule
EEGNet & 0.4090$\pm$0.0122 & 0.3342$\pm$0.0251 & 0.4124$\pm$0.0141 \\
EEGConformer & 0.4559$\pm$0.0125 & 0.3858$\pm$0.0186 & 0.4514$\pm$0.0107 \\
SPaRCNet & 0.4673$\pm$0.0155 & 0.3978$\pm$0.0289 & 0.4729$\pm$0.0133 \\
ContraWR & 0.4887$\pm$0.0078 & 0.4231$\pm$0.0151 & 0.4884$\pm$0.0074 \\
CNN-Transformer & 0.4697$\pm$0.0132 & 0.4017$\pm$0.0168 & 0.4720$\pm$0.0125 \\
FFCL & 0.4673$\pm$0.0158 & 0.3987$\pm$0.0383 & 0.4699$\pm$0.0145 \\
ST-Transformer & 0.4810$\pm$0.0079 & 0.4137$\pm$0.0133 & 0.4795$\pm$0.0096 \\
BIOT & 0.5118$\pm$0.0118 & 0.4476$\pm$0.0254 & 0.5136$\pm$0.0112 \\
LaBraM-Base & 0.5273$\pm$0.0107 & 0.4698$\pm$0.0188 & 0.5288$\pm$0.0102 \\
CBraMod+Linear & 0.5475$\pm$0.0060 & 0.4873$\pm$0.0123 & 0.5463$\pm$0.0167 \\
{\bfseries\boldmath MSCGC-KAN (ours)} & {\bfseries\boldmath 0.6066$\pm$0.0076} & {\bfseries\boldmath 0.5525$\pm$0.0141} & {\bfseries\boldmath 0.6040$\pm$0.0262} \\
\bottomrule
\end{tabular}
}
\end{table}

\subsection{Results on the SEED-VII Dataset}

The SEED-VII results should be interpreted under this subject-dependent protocol. They quantify within-subject-session trial-disjoint performance and are not directly comparable with the FACED results as evidence of cross-subject generalization.

For comparability, we reimplemented EEGNet \cite{ref36}, EEGConformer \cite{ref37}, CSBrain \cite{ref43}, and REVE \cite{ref44} in a unified SEED-VII pipeline. Our EEGNet and EEGConformer results were below the published values. To avoid favoring MSCGC-KAN with lower baselines, the table retains their higher literature results. REVE, CSBrain, CBraMod+Linear, and MSCGC-KAN report five runs with identical preprocessing, splits, optimization/evaluation, seeds, metrics, and checkpoint selection.

Among the controlled rows, MSCGC-KAN and REVE are close. MSCGC-KAN leads in balanced accuracy ($0.3327$) and weighted F1 ($0.3364$), whereas REVE leads in Kappa ($0.2251$). Neither dominates, so the differences are interpreted descriptively. Compared with the linear CBraMod head, MSCGC-KAN improves the three metrics by $0.0203$, $0.0216$, and $0.0174$, respectively. Thus, the structured head improves the same backbone and performs comparably to a strong foundation-model alternative, although absolute seven-class performance remains limited.

\begin{table}[!t]
\renewcommand{\arraystretch}{1.15}
\caption{Performance comparison on SEED-VII (mean$\pm$std). REVE, CSBrain, CBraMod+Linear, and MSCGC-KAN are controlled five-run results. EEGNet and EEGConformer retain higher published reference values because our reproductions were lower.}
\label{tab:seed_compare}
\centering
\resizebox{\columnwidth}{!}{%
\begin{tabular}{lccc}
\toprule
Method & Balanced Accuracy & Kappa & Weighted F1 \\
\midrule
EEGNet & 0.2528$\pm$0.0033 & 0.1285$\pm$0.0044 & 0.2536$\pm$0.0036 \\
EEGConformer & 0.2875$\pm$0.0028 & 0.1747$\pm$0.0030 & 0.2802$\pm$0.0028 \\
CBraMod+Linear & 0.3124$\pm$0.0041 & 0.2007$\pm$0.0052 & 0.3190$\pm$0.0056 \\
CSBrain & 0.3276$\pm$0.0040 & 0.2161$\pm$0.0039 & 0.3305$\pm$0.0031 \\
REVE & 0.3316$\pm$0.0094 & {\bfseries\boldmath 0.2251$\pm$0.0112} & 0.3325$\pm$0.0120 \\
{\bfseries\boldmath MSCGC-KAN (ours)} & {\bfseries\boldmath 0.3327$\pm$0.0056} & 0.2223$\pm$0.0088 & {\bfseries\boldmath 0.3364$\pm$0.0075} \\
\bottomrule
\end{tabular}
}
\end{table}

\subsection{Ablation Experiments}

The FACED ablation in Table~\ref{tab:faced_results} isolates each task-head component. Relative to the CBraMod+Linear baseline, KANLinear improves all metrics, while MCRBlock-GCN provides additional spatiotemporal modeling. Their combination performs best, reaching 60.66\% balanced accuracy, 0.5525 Kappa, and 60.40\% weighted F1-score.

\begin{table}[!t]
\renewcommand{\arraystretch}{1.15}
\caption{Ablation results on the FACED dataset. Values are mean $\pm$ standard deviation across five runs; brackets give the two-sided 95\% confidence interval for the mean.}
\label{tab:faced_results}
\centering
\resizebox{\columnwidth}{!}{%
\begin{tabular}{lccc}
\toprule
Configuration & Balanced Accuracy & Kappa & Weighted F1 \\
\midrule
Baseline (CBraMod+Linear) & \shortstack{$0.5475\pm0.0060$\\\scriptsize $[0.5401,0.5549]$} & \shortstack{$0.4873\pm0.0123$\\\scriptsize $[0.4720,0.5026]$} & \shortstack{$0.5463\pm0.0167$\\\scriptsize $[0.5256,0.5670]$} \\
+KANLinear & \shortstack{$0.5702\pm0.0071$\\\scriptsize $[0.5614,0.5790]$} & \shortstack{$0.5106\pm0.0128$\\\scriptsize $[0.4947,0.5265]$} & \shortstack{$0.5669\pm0.0194$\\\scriptsize $[0.5428,0.5910]$} \\
+MCRBlock-GCN & \shortstack{$0.5553\pm0.0065$\\\scriptsize $[0.5472,0.5634]$} & \shortstack{$0.4944\pm0.0131$\\\scriptsize $[0.4781,0.5107]$} & \shortstack{$0.5513\pm0.0176$\\\scriptsize $[0.5294,0.5732]$} \\
{\bfseries\boldmath MSCGC-KAN (full model)} & \shortstack{{\bfseries\boldmath $0.6066\pm0.0076$}\\\scriptsize $[0.5972,0.6160]$} & \shortstack{{\bfseries\boldmath $0.5525\pm0.0141$}\\\scriptsize $[0.5350,0.5700]$} & \shortstack{{\bfseries\boldmath $0.6040\pm0.0262$}\\\scriptsize $[0.5715,0.6365]$} \\
\bottomrule
\end{tabular}
}
\end{table}

The SEED-VII ablation in Table~\ref{tab:seed_results} differs from the FACED result at the single-module level. KANLinear improves all three metrics relative to the CBraMod+Linear baseline. When MCRBlock-GCN is introduced alone, balanced accuracy increases slightly from 0.3124 to 0.3143, whereas Kappa decreases from 0.2007 to 0.1996 and weighted F1 decreases from 0.3190 to 0.3147. Thus, MCRBlock-GCN alone does not consistently improve all evaluation metrics on SEED-VII. The complete model nevertheless achieves the highest mean values across all three metrics.

\begin{table}[!t]
\renewcommand{\arraystretch}{1.15}
\caption{Ablation results on the SEED-VII dataset. Values are mean $\pm$ standard deviation across five runs; brackets give the two-sided 95\% confidence interval for the mean.}
\label{tab:seed_results}
\centering
\resizebox{\columnwidth}{!}{%
\begin{tabular}{lccc}
\toprule
Configuration & Balanced Accuracy & Kappa & Weighted F1 \\
\midrule
Baseline (CBraMod+Linear) & \shortstack{$0.3124\pm0.0041$\\\scriptsize $[0.3073,0.3175]$} & \shortstack{$0.2007\pm0.0052$\\\scriptsize $[0.1942,0.2072]$} & \shortstack{$0.3190\pm0.0056$\\\scriptsize $[0.3120,0.3260]$} \\
+KANLinear & \shortstack{$0.3205\pm0.0048$\\\scriptsize $[0.3145,0.3265]$} & \shortstack{$0.2094\pm0.0058$\\\scriptsize $[0.2022,0.2166]$} & \shortstack{$0.3226\pm0.0061$\\\scriptsize $[0.3150,0.3302]$} \\
+MCRBlock-GCN & \shortstack{$0.3143\pm0.0045$\\\scriptsize $[0.3087,0.3199]$} & \shortstack{$0.1996\pm0.0054$\\\scriptsize $[0.1929,0.2063]$} & \shortstack{$0.3147\pm0.0059$\\\scriptsize $[0.3074,0.3220]$} \\
{\bfseries\boldmath MSCGC-KAN (full model)} & \shortstack{{\bfseries\boldmath $0.3327\pm0.0056$}\\\scriptsize $[0.3257,0.3397]$} & \shortstack{{\bfseries\boldmath $0.2223\pm0.0088$}\\\scriptsize $[0.2114,0.2332]$} & \shortstack{{\bfseries\boldmath $0.3364\pm0.0075$}\\\scriptsize $[0.3271,0.3457]$} \\
\bottomrule
\end{tabular}
}
\end{table}

Across both datasets, KANLinear provides the larger standalone gain. MCRBlock-GCN provides a modest standalone improvement on FACED but not a consistent improvement on SEED-VII. Within the complete architecture, MCRBlock-GCN reorganizes the representation through multi-scale temporal and task-driven inter-channel operations, while KANLinear subsequently maps it into an analytic nonlinear feature space. The combined configuration achieves the highest mean values in both ablations. Together with the stage-wise feature-space visualization, this result is consistent with complementary roles under the present protocol. It does not, however, establish mechanistic synergy between the modules or imply a guaranteed standalone benefit from MCRBlock-GCN.

\subsection{Visualization}

\subsubsection{Confusion Matrix Analysis}

\begin{figure*}[!t]
\centering
\includegraphics[width=0.82\textwidth]{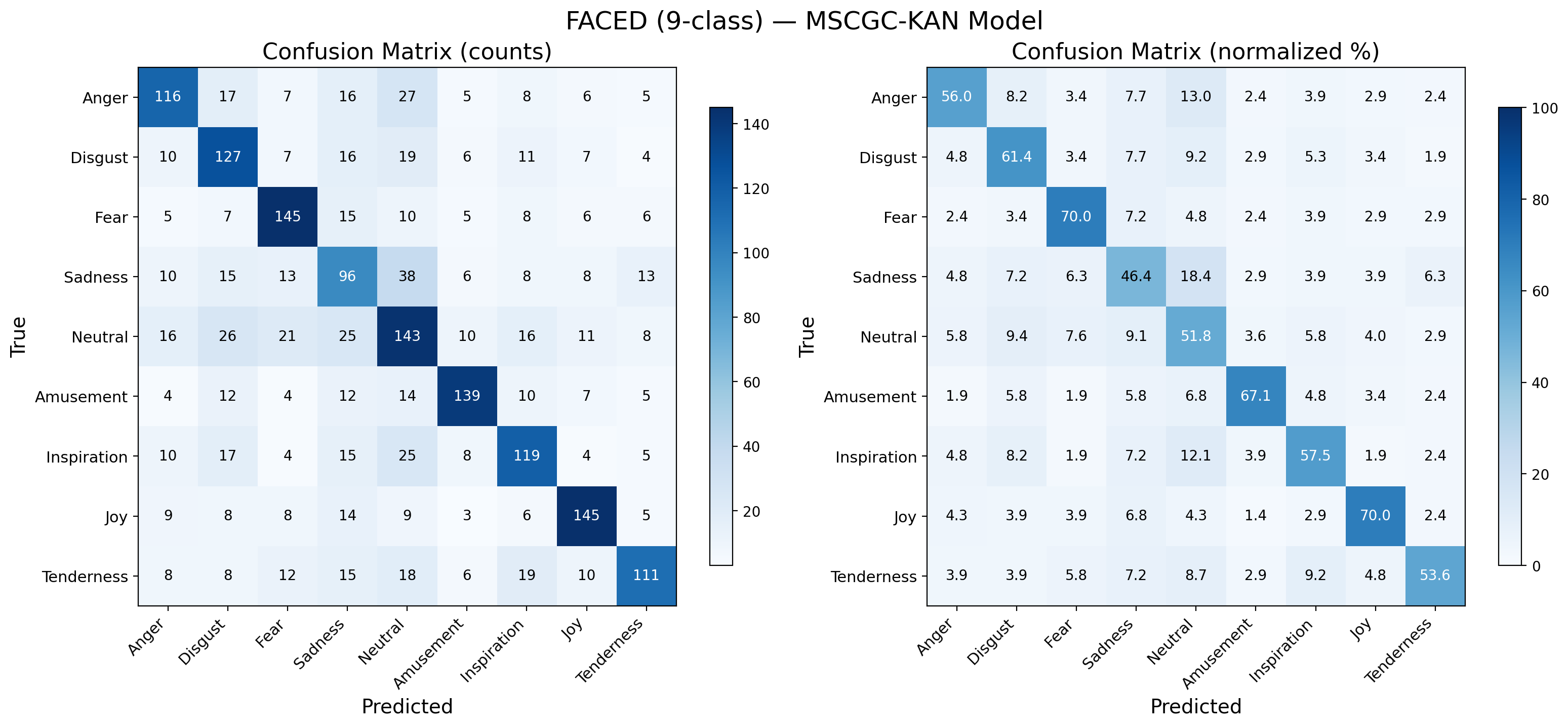}
\caption{Confusion matrix on FACED. Major confusions are concentrated among emotionally similar classes, reflecting the local similarity structure of affective space.}
\label{fig:confusion_faced}
\end{figure*}

\begin{figure*}[!t]
\centering
\includegraphics[width=0.82\textwidth]{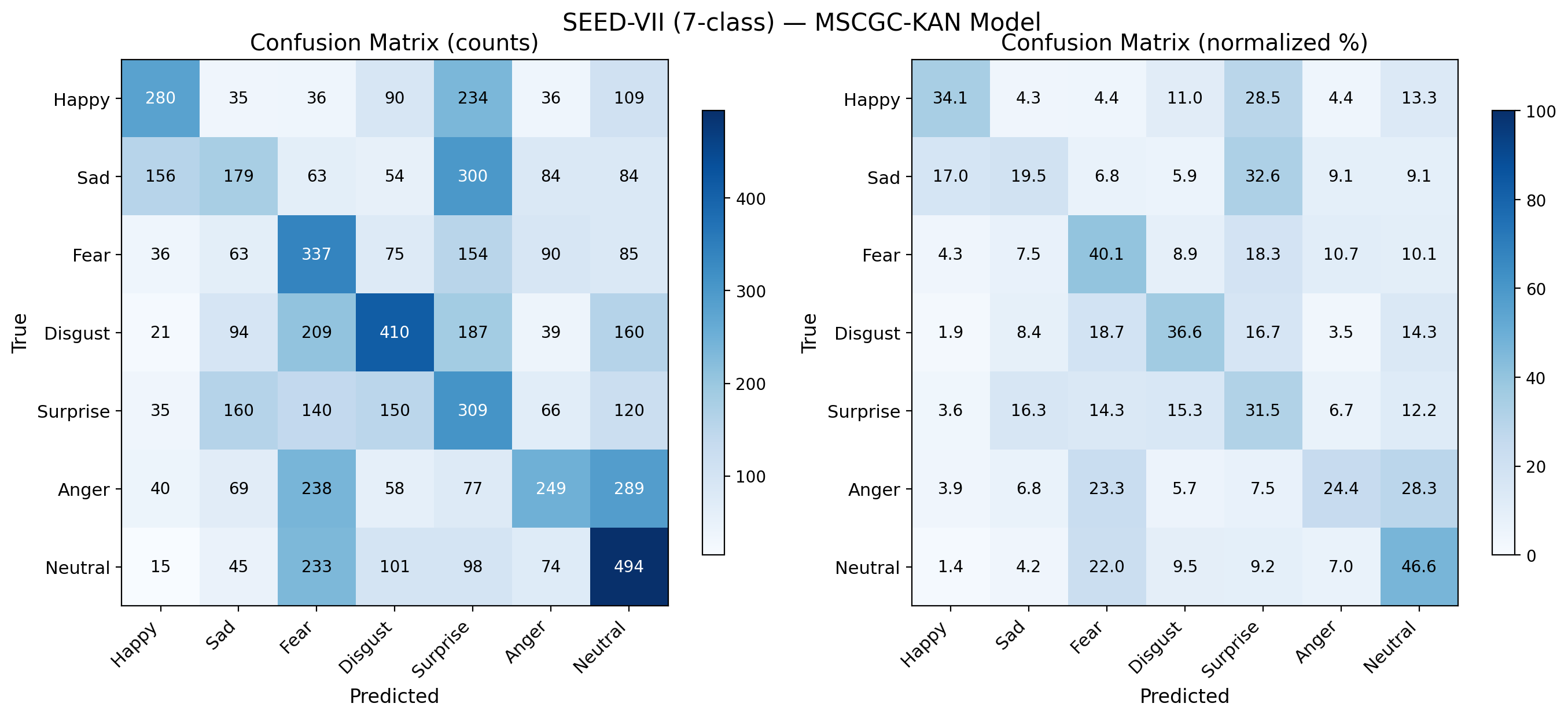}
\caption{Confusion matrix on SEED-VII. Correct predictions are concentrated along the diagonal for several categories, while substantial off-diagonal errors remain.}
\label{fig:confusion_seed}
\end{figure*}

Figs.~\ref{fig:confusion_faced} and \ref{fig:confusion_seed} present the confusion matrices on the two datasets. FACED shows clearer diagonal concentration, whereas SEED-VII retains substantial off-diagonal errors and low recalls for several categories. On FACED, the highest recalls appear for Fear and Joy, both reaching 70.0\%, whereas Sadness is more difficult to distinguish, with a recall of 46.4\%. The most visible off-diagonal errors mainly occur around neutral or semantically adjacent categories: Sadness is misclassified as Neutral by 18.4\%, Anger is misclassified as Neutral by 13.0\%, Inspiration is misclassified as Neutral by 12.1\%, and Tenderness is misclassified as Inspiration by 9.2\%. This pattern is reasonable for FACED because the dataset contains fine-grained positive and negative categories as well as a neutral class, so several classes lie close to each other in affective semantics \cite{ref25}.

The SEED-VII confusion matrix exhibits stronger category overlap. Happy has a recall of 34.1\% and is frequently predicted as Surprise by 28.5\%. Sad is predicted as Surprise by 32.6\%, while Surprise is also confused with Sad, Disgust, and Fear by 16.3\%, 15.3\%, and 14.3\%, respectively. Anger is predicted as Neutral and Fear by 28.3\% and 23.3\%, and Neutral is predicted as Fear by 22.0\%. These quantitative patterns show that the main errors are not uniformly distributed across all classes, but concentrate on several specific emotion pairs.

\subsubsection{Feature Space Visualization}

\begin{figure*}[!t]
\centering
\includegraphics[width=0.98\textwidth]{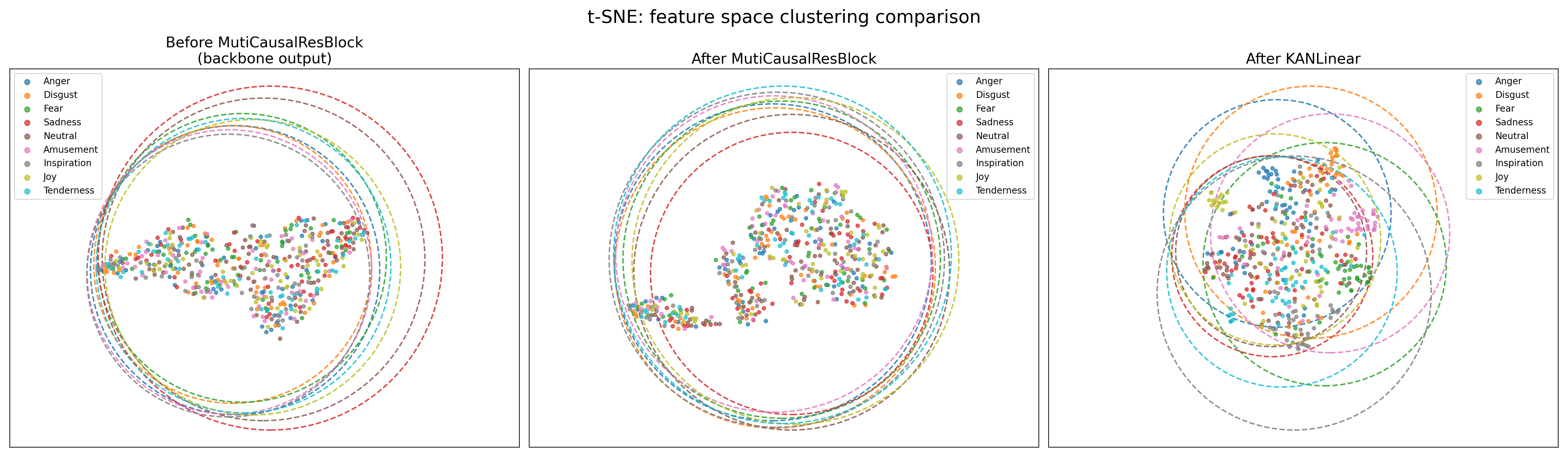}
\caption{t-SNE visualization of stage-wise feature spaces on FACED. The three stages qualitatively show changes in inter-class separation and intra-class compactness.}
\label{fig:tsne_faced}
\end{figure*}

\begin{figure*}[!t]
\centering
\includegraphics[width=0.98\textwidth]{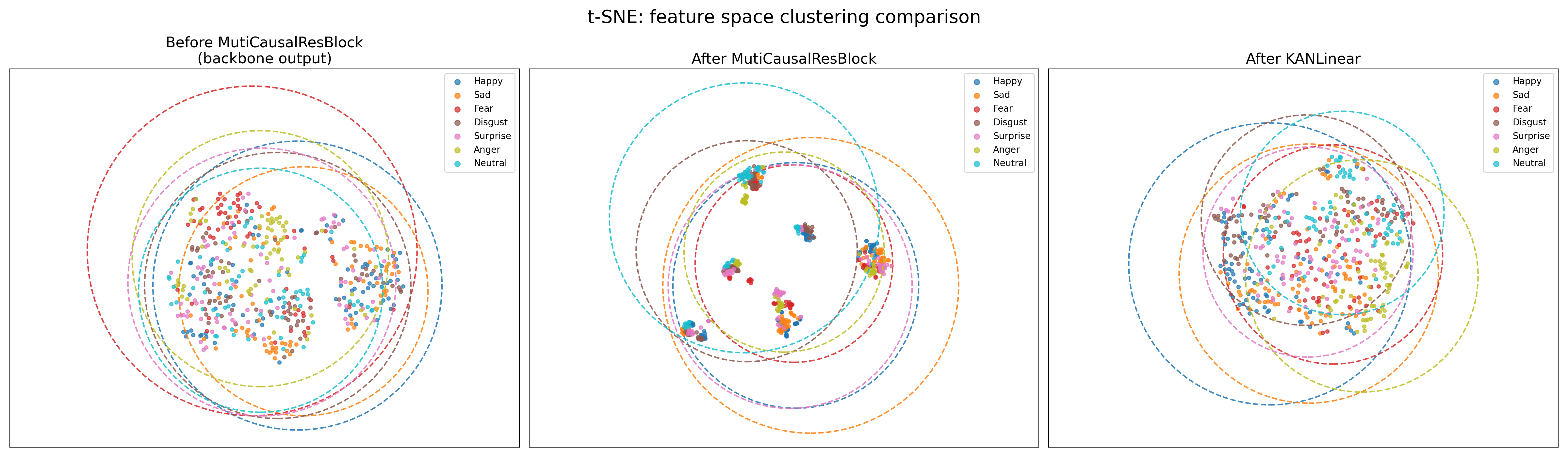}
\caption{t-SNE visualization of stage-wise feature spaces on SEED-VII. The MCRBlock-GCN output retains overlap among several emotion categories, whereas the KANLinear-mapped representation shows more compact within-class grouping and clearer separation among several categories. This visualization provides qualitative evidence of feature-space refinement.}
\label{fig:tsne_seed}
\end{figure*}

To examine how the task head transforms the pretrained representation, Figs.~\ref{fig:tsne_faced} and \ref{fig:tsne_seed} show t-SNE projections of the backbone output, the MCRBlock-GCN representation, and the KANLinear-mapped representation. On SEED-VII, the representations after MCRBlock-GCN retain considerable overlap among several emotion categories. After KANLinear mapping, the projected features show more compact within-class distributions and clearer separation among several categories. This stage-wise change is consistent with KANLinear further refining the graph-enhanced representation into a more discriminative feature space. Because t-SNE is a nonlinear low-dimensional visualization, this qualitative pattern is not treated as definitive mechanistic evidence of module synergy.

\subsubsection{Ablation Experiment Visualization}

\begin{figure*}[!t]
\centering
\begin{minipage}[t]{0.48\textwidth}
\centering
\includegraphics[width=\linewidth]{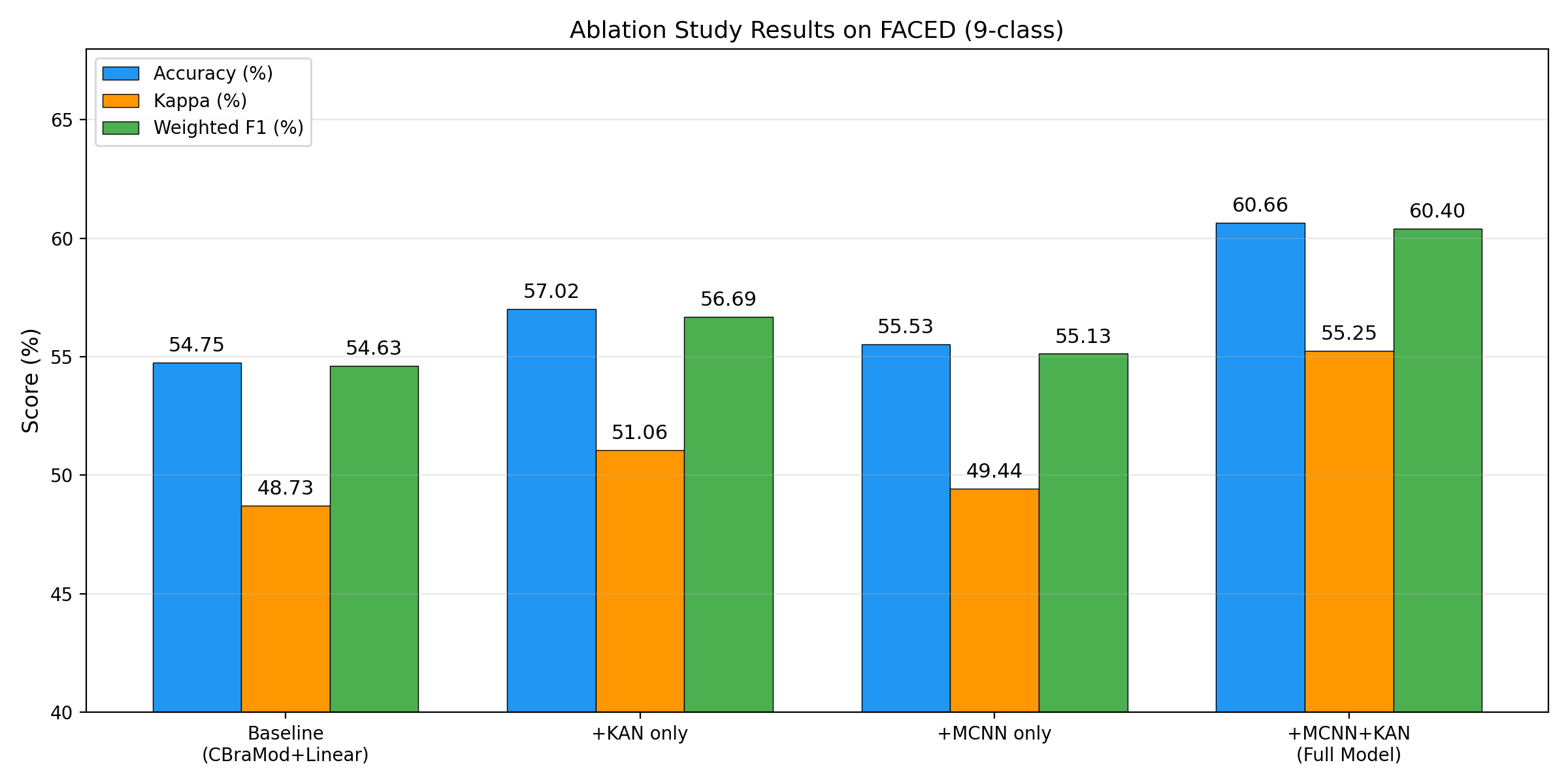}
\end{minipage}\hfill
\begin{minipage}[t]{0.48\textwidth}
\centering
\includegraphics[width=\linewidth]{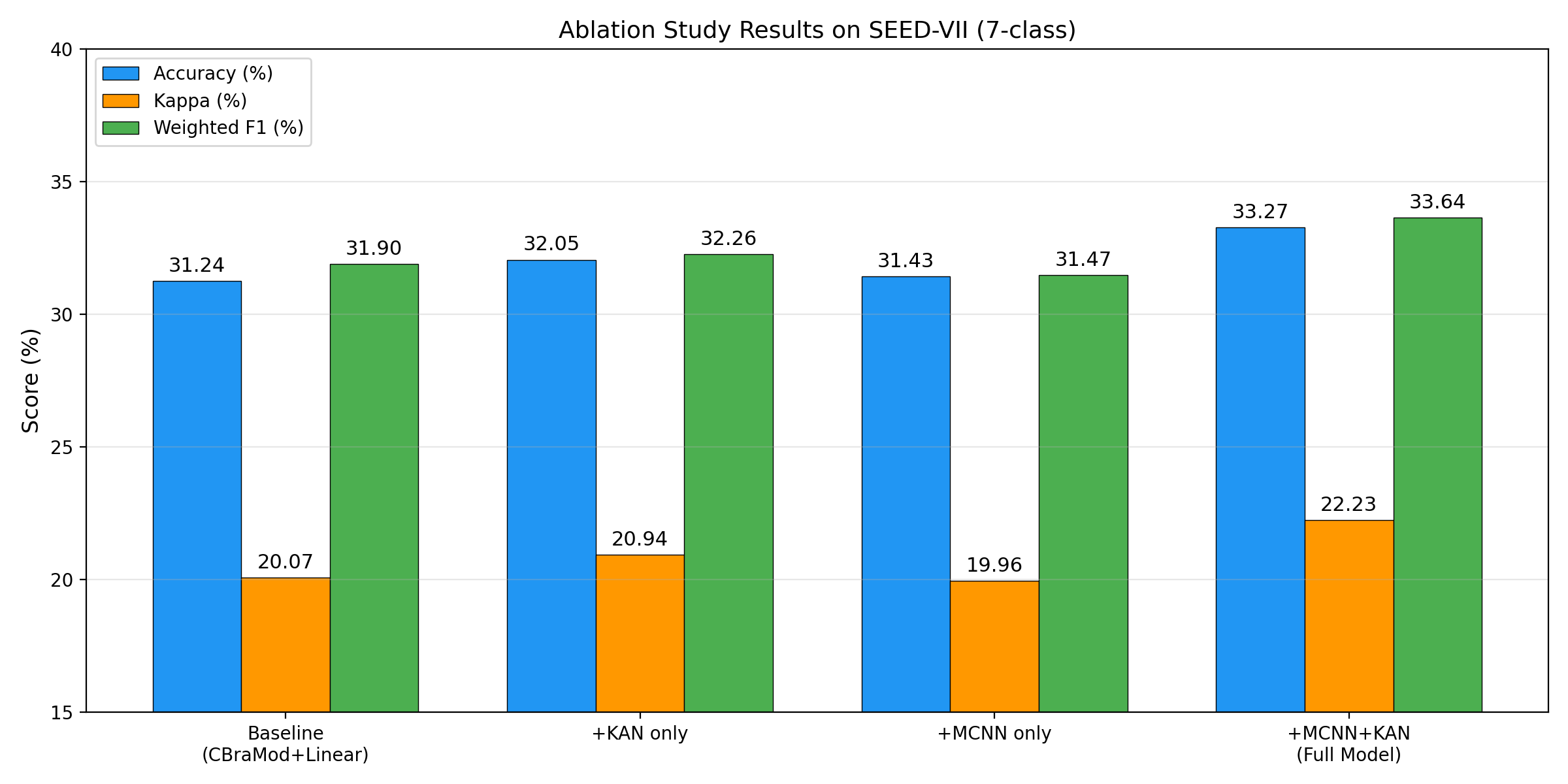}
\end{minipage}
\caption{Visualization of ablation results on FACED and SEED-VII. The left panel corresponds to FACED and the right panel corresponds to SEED-VII. The grouped bars show the dataset-dependent standalone behavior of MCRBlock-GCN and the highest mean performance of the complete model on both datasets.}
\label{fig:ablation_bars}
\end{figure*}

To present the dataset-specific contributions of the two modules more clearly, Fig.~\ref{fig:ablation_bars} visualizes the FACED and SEED-VII ablation results as grouped bar charts. On FACED, both individual modules improve all three metrics. On SEED-VII, KANLinear improves all three metrics, whereas MCRBlock-GCN alone slightly improves balanced accuracy but reduces Kappa and weighted F1. The complete MSCGC-KAN model obtains the highest mean values among the ablation configurations on both datasets.

\subsubsection{Learned Task-driven Channel-dependency Patterns}

To avoid excessive repetition of model-level visualizations, representative visualizations are presented on FACED in the main text, while the corresponding SEED-VII visualizations are provided in the Supplementary Material. Fig.~\ref{fig:adjacency_faced} visualizes the learned channel-dependency matrix, task-driven channel-dependency graph, and ranking of channels by aggregate dependency weight. Channels with strong edges or high aggregate weights are interpreted as having prominent roles in the model's discriminative dependency structure, not as physiological functional hubs. The intra-prefrontal, prefrontal--temporal, and cross-hemispheric edges likewise indicate task-relevant dependencies learned under the classification objective rather than direct evidence of neural functional coupling.

\begin{figure*}[!t]
\centering
\includegraphics[width=0.98\textwidth]{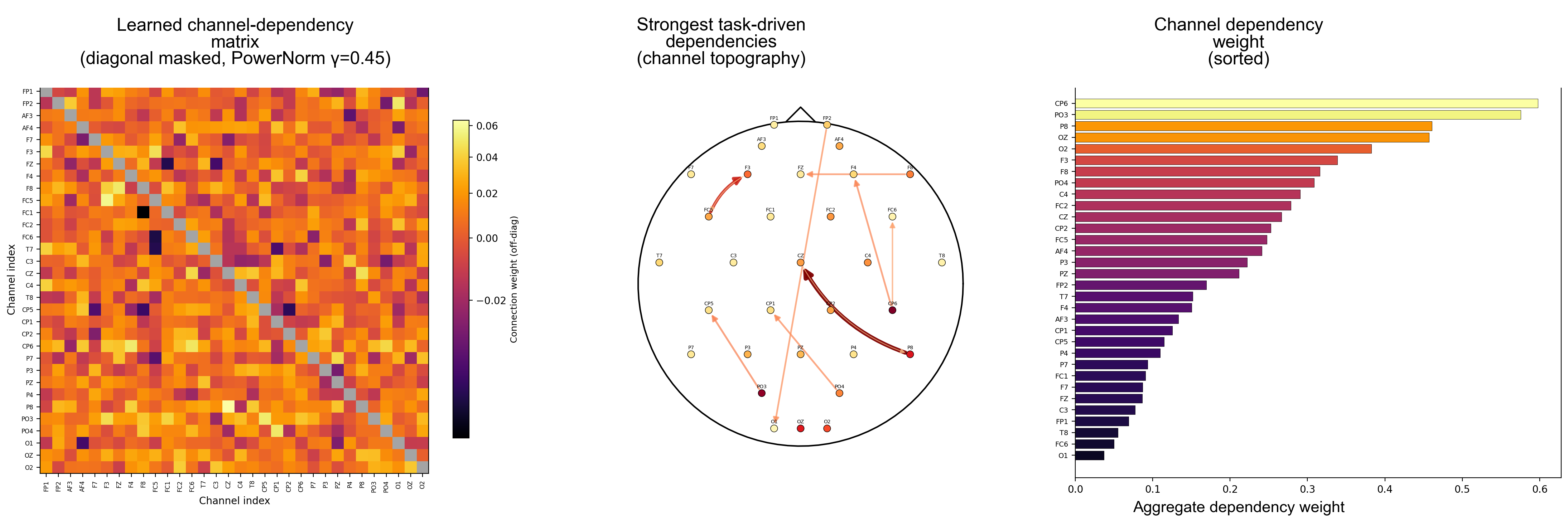}
\caption{Visualization of the learned channel-dependency matrix on FACED, including the dependency-weight heat map, the task-driven channel-dependency graph, and the ranking of channels by aggregate dependency weight. These patterns characterize model behavior under the classification objective.}
\label{fig:adjacency_faced}
\end{figure*}

\subsubsection{Spatial Activation Topography}

\begin{figure*}[!t]
\centering
\includegraphics[width=0.98\textwidth]{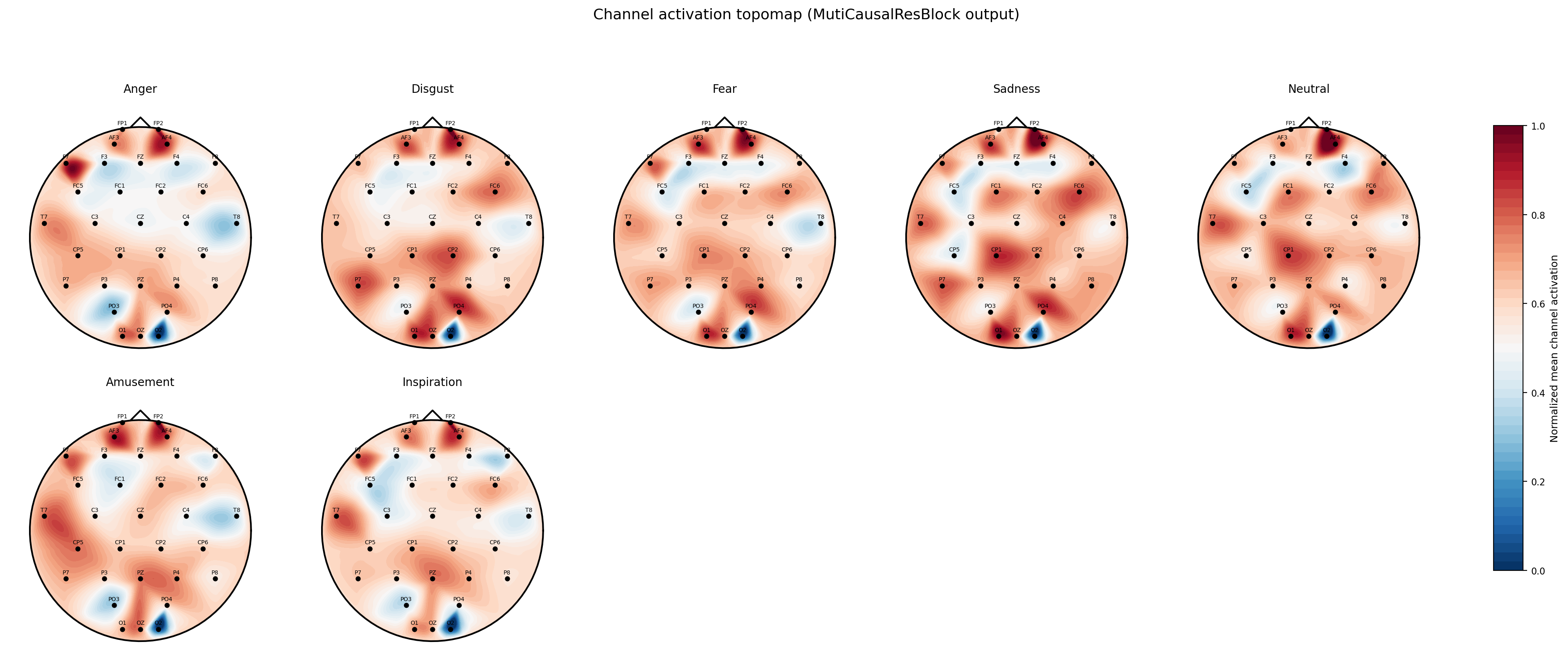}
\caption{Model-derived scalp topography on FACED. The maps visualize class-dependent channel responses used by the classifier.}
\label{fig:topomap_faced}
\end{figure*}

Fig.~\ref{fig:topomap_faced} presents model-derived channel-response patterns from the perspective of scalp topography. For FACED, the visualizations show class-dependent differences in model responses over prefrontal and temporal channels, while Neutral produces a comparatively even pattern. These maps characterize spatial features used by the classifier and are not treated as direct evidence of emotion-related cortical activation. Because FACED preprocessing uses FP1/FP2 as proxy EOG channels during ICA, residual ocular activity cannot be excluded as a contributor to prefrontal responses. We therefore do not attribute their prominence specifically to emotion-related neural activity.

\subsubsection{Temporal Attribution Analysis}

\begin{figure*}[!t]
\centering
\includegraphics[width=0.98\textwidth]{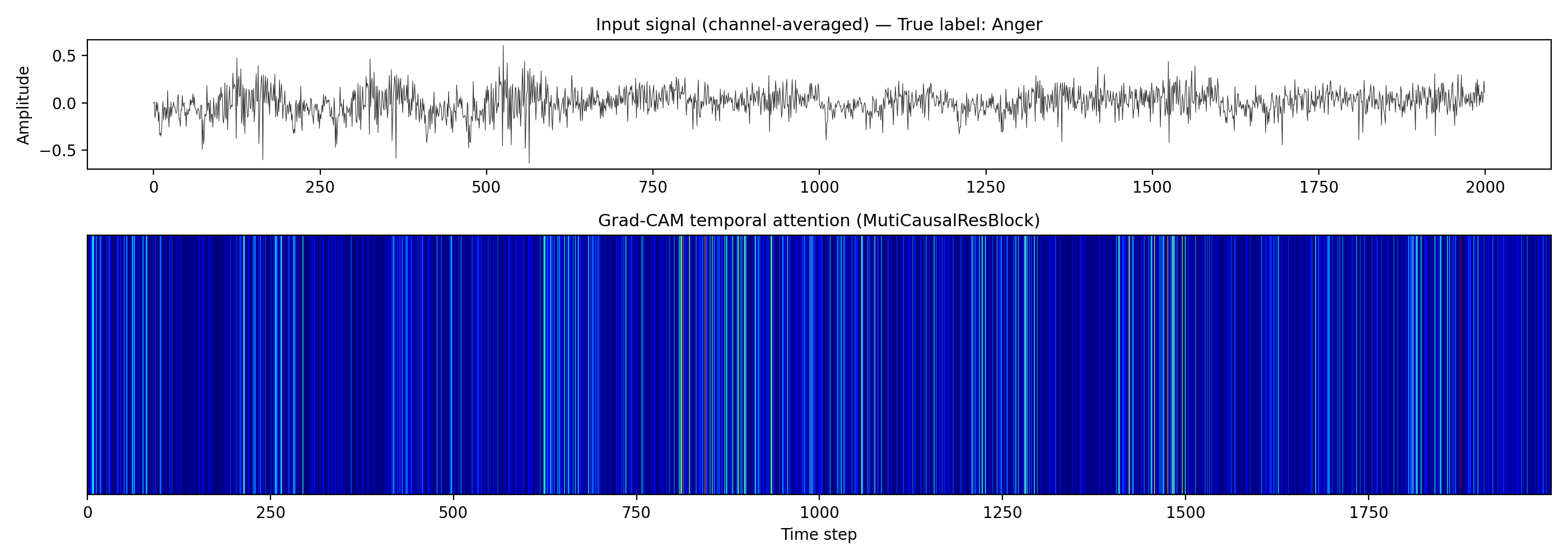}
\caption{One-dimensional Grad-CAM temporal attribution on FACED, computed from the MCRBlock-GCN output before flattening. Higher values indicate larger positive contributions to the selected class score.}
\label{fig:gradcam_faced}
\end{figure*}

Grad-CAM is applied to the MCRBlock-GCN output $\boldsymbol{H}$ rather than to Transformer attention maps. For one sample, $\boldsymbol{H}\in\mathbb{R}^{C\times T}$, where $C$ indexes channel-aligned feature maps and $T=SD$ indexes the ordered positions of the reshaped CBraMod representation before flattening and classification. For target-class score $s_{\ell}$, the channel weight and temporal attribution are $\alpha_i^{(\ell)}=T^{-1}\sum_{t=1}^{T}\partial s_{\ell}/\partial H_{i,t}$ and $g_t^{(\ell)}=\operatorname{ReLU}(\sum_{i=1}^{C}\alpha_i^{(\ell)}H_{i,t})$, respectively, followed by max normalization. Fig.~\ref{fig:gradcam_faced} shows this one-dimensional gradient-weighted temporal attribution profile on FACED. Higher values indicate temporal positions with larger positive contributions to the selected class score. They are not Transformer attention weights or validated physiological temporal markers.

\subsubsection{KANLinear Basis-function Visualization}

\begin{figure*}[!t]
\centering
\begin{minipage}[t]{0.48\textwidth}
\centering
\includegraphics[width=\linewidth]{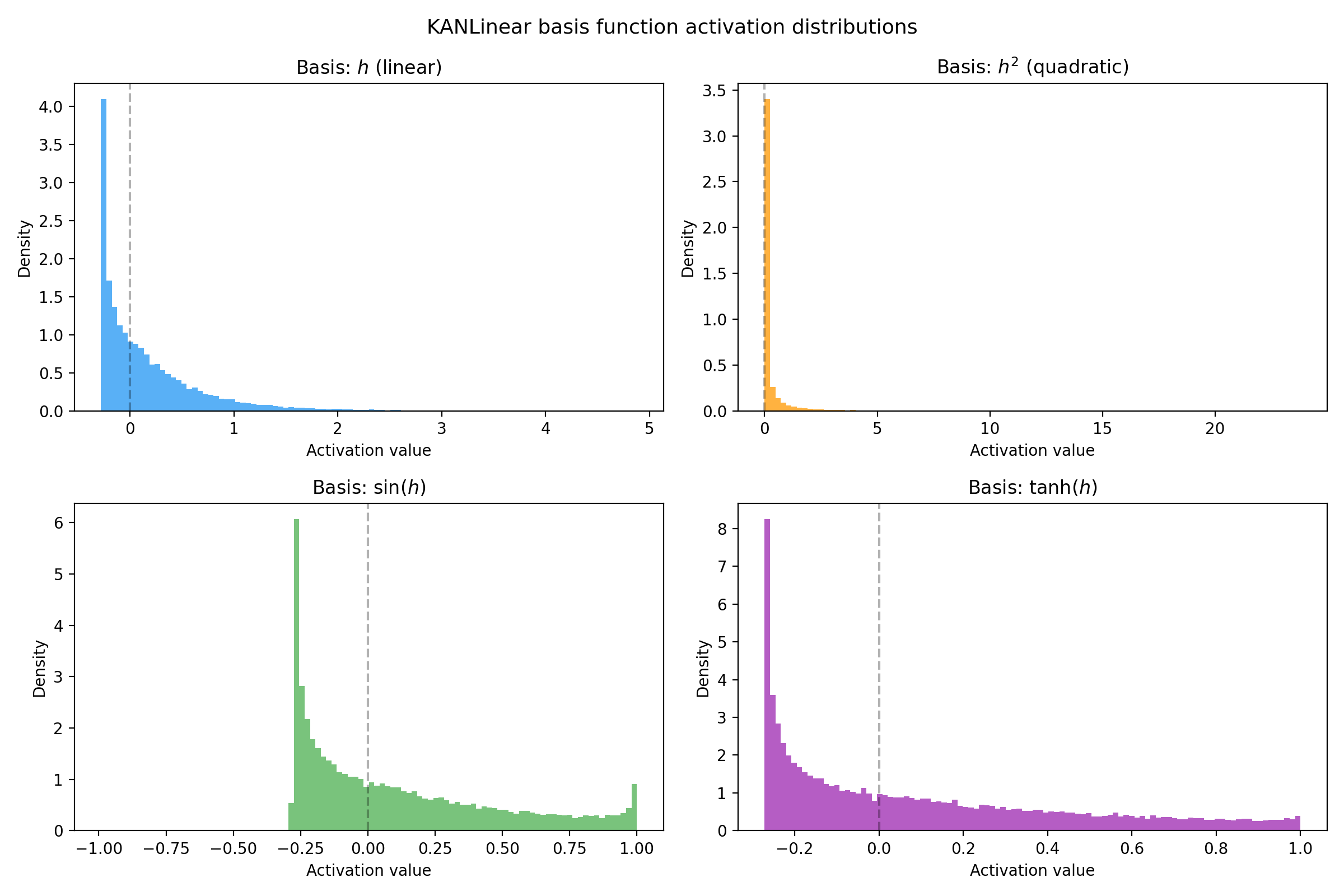}
\end{minipage}\hfill
\begin{minipage}[t]{0.48\textwidth}
\centering
\includegraphics[width=\linewidth]{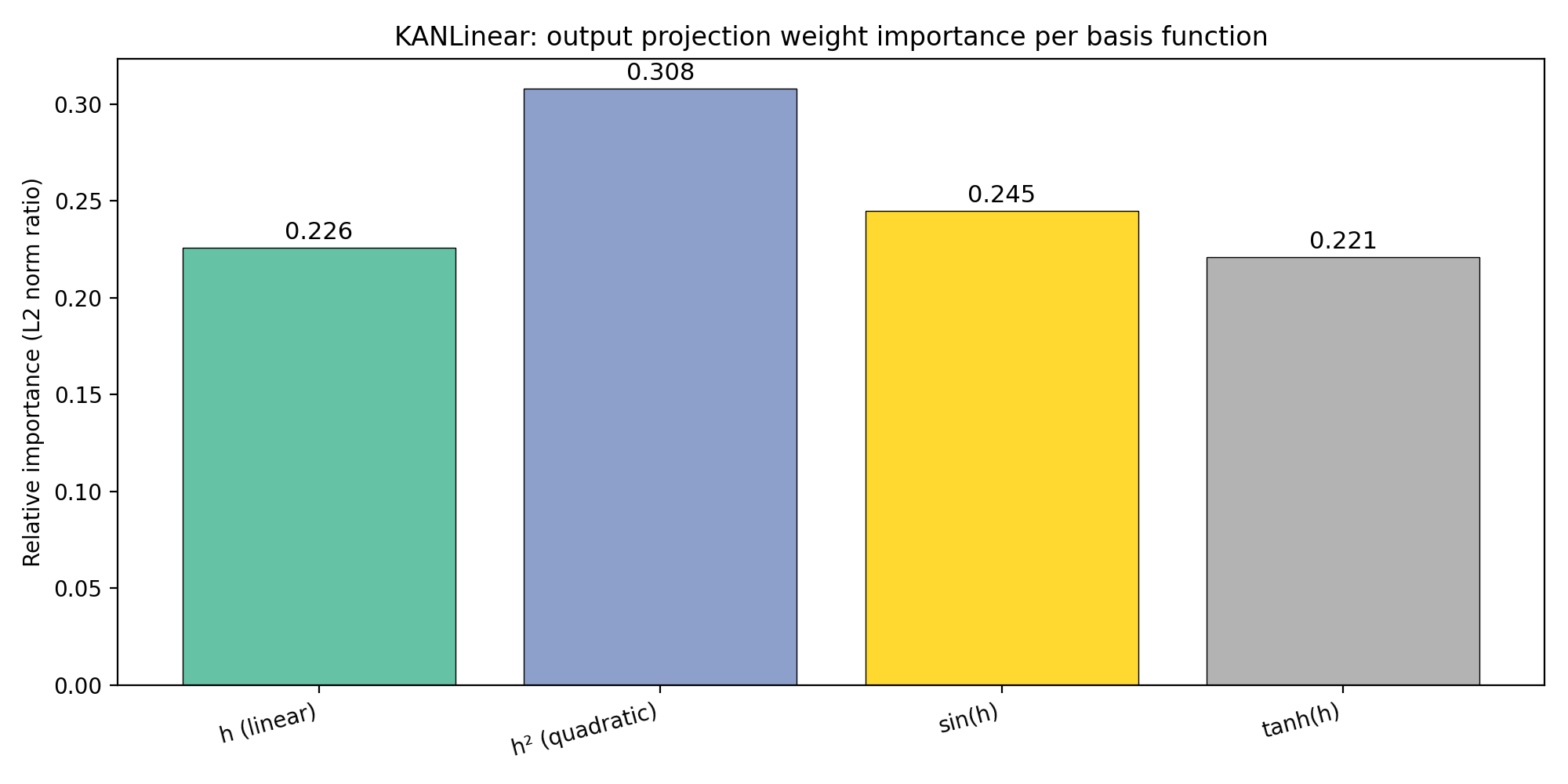}
\end{minipage}
\caption{Basis-response distributions and projection-weight importance of KANLinear on FACED. The linear term and multiple nonlinear basis functions jointly contribute to emotion-discriminative mapping.}
\label{fig:kan_faced}
\end{figure*}

Fig.~\ref{fig:kan_faced} presents the basis-response distributions and output-projection weight importance of KANLinear on FACED. The linear, quadratic, sinusoidal, and hyperbolic tangent components show different response ranges and different projection-weight contributions. The visualization shows that multiple analytic basis functions participate in the final feature mapping rather than only one dominant component.

\section{Discussion}

This paper focuses on the problem of downstream task-head design for EEG emotion recognition under the foundation-model fine-tuning setting. To address the limited spatiotemporal modeling and nonlinear discrimination ability of a simple linear head, MSCGC-KAN introduces multi-scale causal graph convolution and KANLinear, a KAN-inspired analytic-basis mapping module, after the pre-trained CBraMod backbone. The experimental results on FACED and SEED-VII show consistent improvements over the CBraMod+Linear baseline, with balanced-accuracy gains of 5.91 and 2.03 percentage points, respectively. These improvements are protocol-specific. FACED evaluates subject-disjoint generalization, whereas SEED-VII evaluates trial-disjoint classification with subject and session overlap. They should therefore not be interpreted as equivalent evidence of cross-subject generalization. These results indicate that the discriminative ability of pre-trained EEG representations can be further improved by a structured task-specific head without redesigning the backbone.

Relative to CBraMod+Linear, MSCGC-KAN reduces total parameters by 68.37\% and 83.75\%, FLOPs/sample by 5.16\% and 6.03\%, and batch-one FP32 peak inference memory by 66.62\% and 82.40\% on FACED and SEED-VII, respectively (memory measured on an RTX 2060). Both configurations require about 0.8 min/epoch on FACED and 1.6 min/epoch on SEED-VII for 30 epochs. These results support describing MSCGC-KAN as compact, lightweight, and parameter-efficient relative to CBraMod+Linear, with lower resource use and comparable training time.

Because CBraMod applies criss-cross attention to the complete input segment, the backbone itself is not constrained to operate causally. Causal padding within the downstream temporal module therefore does not make the complete pipeline an online model. We interpret it only as a directional local inductive bias and do not treat it as a necessary condition for the offline classification task.

The results are also related to previous EEG emotion recognition studies that emphasize temporal, spatial, and graph-based modeling. Dynamic graph convolution methods have shown that adaptive channel-relation learning is useful for EEG emotion recognition \cite{ref13}. Hierarchical spatial-temporal modeling has also been shown to help organize affective EEG features from local electrode patterns to broader brain-region representations \cite{ref31}. Our learned channel-dependency and topographic patterns show qualitative consistency with these studies: prefrontal and temporal channels receive prominent model responses, and the learned graph contains cross-region and cross-hemispheric task-driven dependencies. This resemblance does not confirm that the model has recovered physiological functional connectivity. Task-driven graph learning and physiological connectivity estimation are different objectives, and the present adjacency matrix should primarily be interpreted as evidence about model behavior and discriminative representation learning.

Compared with previous task-specific EEG emotion models, the main difference of MSCGC-KAN is that the proposed modules are used as a downstream adaptation head on top of a pre-trained EEG backbone. Compared with the CBraMod+Linear baseline, MCRBlock-GCN explicitly reorganizes the feature space through multi-scale temporal convolution and learnable graph convolution, while KANLinear further expands the representation using multiple analytic basis functions. The ablation results show dataset-dependent standalone behavior. KANLinear alone improves all three metrics on both datasets. MCRBlock-GCN alone improves all three metrics on FACED, but on SEED-VII it slightly increases balanced accuracy while reducing Kappa and weighted F1. The complete configuration nevertheless achieves the highest mean values on both datasets. The stage-wise t-SNE analysis provides qualitative context for this pattern. On SEED-VII, the representation after MCRBlock-GCN retains substantial overlap among several emotion categories, whereas KANLinear mapping produces more compact within-class distributions and clearer separation among several categories. This pattern is consistent with KANLinear refining the graph-enhanced representation and may help explain why MCRBlock-GCN contributes within the complete architecture despite its weaker standalone result. However, t-SNE cannot establish a causal interaction between modules. We therefore describe the observed relationship as a complementary interaction under the present protocol rather than definitive mechanistic synergy.

On FACED, some residual errors involve semantically related classes, including Sadness and Neutral. On SEED-VII, however, the prominent Happy-to-Surprise and Sad-to-Surprise errors are not consistently explained by valence/arousal proximity because Happy and Sad have opposite valence. The circumplex model therefore does not adequately account for these directional confusions. We interpret them more conservatively as unresolved category overlap and a limitation of the model in this regime. The SEED-VII results show improvement over the directly evaluated CBraMod+Linear baseline, but the modest gains coexist with limited absolute discrimination and substantial class confusion. The feature-space, channel-dependency, topographic, temporal-saliency, and KANLinear basis-response visualizations provide qualitative model-level views of spatial responses, temporal attribution, and internal nonlinear mapping. They help inspect the model's classification behavior but do not establish neurophysiological mechanisms or validated functional connectivity. Thus, they complement the performance results by making downstream adaptation more transparent at the model level.

\section{Conclusion}

This paper investigated downstream classification-head design for EEG foundation models and proposed MSCGC-KAN, an emotion recognition framework built around multi-scale causal graph convolution and KANLinear-based analytic-basis mapping. Using CBraMod as the pre-trained backbone, the method strengthens multi-scale temporal modeling and task-driven inter-channel dependency learning through MCRBlock-GCN, and then improves nonlinear discrimination through KANLinear. Differential learning rates and stabilization strategies further support effective fine-tuning without redesigning or re-pretraining the backbone.

The main strength of MSCGC-KAN is that it offers a compact way to adapt generic EEG representations to fine-grained affective pattern recognition while supporting qualitative inspection of the resulting model behavior. Experiments on FACED and SEED-VII show consistent gains over the CBraMod linear baseline, with balanced-accuracy improvements of 5.91 and 2.03 percentage points, respectively. These gains were obtained under different evaluation settings and support effectiveness within each protocol rather than an equivalent claim of cross-subject generalization. These results suggest that researchers and practitioners working with pre-trained EEG models can benefit from paying closer attention to the structure of the task head, rather than treating downstream classification as a simple linear projection. The visualization analyses provide qualitative model-level insight into learned channel dependencies, spatial responses, temporal saliency, and nonlinear mapping behavior. They make the adaptation process more transparent without implying neurophysiological validation.

Several weaknesses remain. The experiments are limited to two public emotion datasets, and broader validation across more datasets, subjects, acquisition protocols, and downstream EEG tasks is needed. Systematic sensitivity analysis of the four settings remains future work. The present SEED-VII protocol is subject-dependent and therefore does not establish unseen-subject generalization. A subject-disjoint SEED-VII evaluation is needed to assess cross-subject transfer and is an important direction for future work. Each directly evaluated configuration was run five times using different random seeds; the mean and standard deviation quantify predictive variability, and the small deviations indicate limited initialization sensitivity. This predictive stability does not establish edge-wise adjacency consistency across seeds, subjects, or evaluation partitions. Future work should quantify such consistency and compare the learned graph with fixed anatomical, electrode-distance-based, correlation-based, and sparsity-constrained alternatives. The adjacency remains task-driven rather than physiological. Likewise, the qualitative channel-dependency, topographic, temporal-saliency, and basis-response visualizations lack cross-seed or partition consistency analysis and randomization validation. Systematic stability and randomization tests are needed to evaluate explanation reliability. Future work will test the proposed structured head with REVE and other EEG foundation-model representations to disentangle backbone and task-head contributions under broader subject-disjoint and efficiency-controlled protocols. Future work will also address adaptive structure search, parameterized basis selection, and online emotion recognition. Incorporating hemispheric asymmetry attention, cross-patch temporal attention, and prototype-based contrastive constraints may further introduce neuroscientific priors and improve the usefulness of foundation-model fine-tuning for broader brain--computer interface applications.

\section*{Data availability}

The data analyzed in this study are the public FACED and SEED-VII datasets cited in the manuscript, which are available from their original providers under their access policies; no new dataset was generated or redistributed.

\bibliographystyle{elsarticle-num}
\bibliography{refs_black}

\end{document}